\documentclass{article}
\usepackage{ijcai19}

\usepackage{natbib}
\usepackage{hyperref}
\usepackage[normalem]{ulem}
\usepackage{url}
\usepackage[T1]{fontenc}

\usepackage{amsfonts}
\usepackage{booktabs}
\usepackage{amssymb}
\usepackage{amsthm}
\usepackage{dsfont}
\usepackage{stmaryrd}
\usepackage{caption}
\usepackage{tikz}
\usetikzlibrary{positioning,calc}

\def\Z{\mathbb{Z}}
\def\R{\mathbb{R}}

\def\ii{\mathrm{ii}}

\def\std{\mathrm{std}}

\def\mean{\mathrm{mean}}

\def\mean{\mathrm{mean}}
\def\recerror{\mathrm{rec\_error}}

\def\D{\mathcal{D}}
\def\E{\mathcal{E}}
\def\Z{\mathcal{Z}}

\def\X{\mathbf{X}}
\def\Y{\mathbf{Y}}

\def\x{\mathrm{x}}

\def\D{\mathcal{D}}

\def\CW{\mathrm{cw}}

\def\1F1{\mbox{$_{1}{F}_{\!1}$}}

\usepackage{color}

\usepackage{todonotes}

\newtheorem*{theorem*}{Theorem}

\usepackage{algorithmic}
\usepackage{algorithm}

\usepackage{color}
\usepackage{todonotes}

\newcommand{\ola}[1]{\textcolor{orange}{#1}}

\title{Non-linear ICA based on Cramer-Wold metric}

\author{
Przemys\l{}aw Spurek \and
Aleksandra Nowak \and
Jacek Tabor \and 
\L{}ukasz Maziarka \and
Stanis\l{}aw Jastrz\k{e}bski \\
\affiliations
Jagiellonian University\\
\emails
przemyslaw.spurek@uj.edu.pl
}

\def\CWICA{CW-ICA}

\begin{document}

\maketitle

\begin{abstract}
Non-linear source separation is a challenging open problem with many applications. We extend a recently proposed Adversarial Non-linear ICA (ANICA) model, and introduce Cramer-Wold ICA (\CWICA{}). In contrast to ANICA we use a simple, closed--form optimization target instead of a discriminator--based independence measure. Our results show that \CWICA{} achieves comparable results to ANICA, while foregoing the need for adversarial training. 
\end{abstract}

\section{Introduction}

Linear Independent Components Analysis (ICA) has become an important data analysis technique. For example, it is routinely used for blind source separation in a wide range of signals. The objective of ICA is to identify a \emph{linear} transformation such that after the projection  the components  of the dataset are independent. More formally, the aim is to find an \emph{unmixing matrix}~$W$ that transforms the observed data $X=(\x_1,\ldots,\x_n)^T$ into maximally independent components $S = WX =(s_1,\ldots,s_n)$ with respect to some measure of independence. Commonly the independence is approximated using a measure of nongaussianity (e.g. kurtosis \cite{hyvarinen1999fast,bell1995information}).

An obvious drawback of ICA is the restriction to linear transformations. Unfortunately, in many practical applications this linearity assumption does not hold, which motivates research into Nonlinear ICA (NICA)~\citep{NIPS2016_6395,pmlr-v70-hirayama17a}. 

One of the key challenges in developing nonlinear variant of ICA is devising an efficient measure of independence. The currently  most popular approach is to constrain the transformation so that independence can be efficiently
estimated~\citep{tan2001nonlinear,almeida2003misep,almeida2004linear,dinh2014nice,zhang2008minimal}. 
Another approach is to \emph{learn} the independence measure. This can be achieved using Generative Adversarial Networks (GANs) \citep{goodfellow2014generative}. In \citep{brakel2017learning} (ANICA - Adversarial Non-linear ICA) authors demonstrate efficacy of using GAN for learning an  independence measure. They show that GAN based independency measure combined with an autoencoder architecture can be used to solve nonlinear blind source separation problems.  

Unfortunately, the use of adversarial training in ANICA comes at the cost of added instability, as also noted by the authors. Our main contribution is \emph{developing an effective independence measure that does not require adversarial training, and matches ANICA performance.} In other words, we found that the adversarial training is not the key contributor to the efficacy of ANICA, and based on this insight we developed a simpler, closed-form independence measure. We demonstrate its efficacy on standard blind source separation problems. 

This paper is structured as follows. We start by discussing related work in Section \ref{sec:rw}. In Section \ref{sec:theo} we describe the key contribution: the independence measure based on \emph{Cramer-Wold metric}. ICA based on the introduced independence measure is described in Sec. \ref{sec:alg}. Finally, we report experimental results in Section \ref{se:ex}.

\section{Related work}
\label{sec:rw}

The fundamental problem in solving NICA is that the solution is in principle non-identifiable. Without any constraints on the space of the mixing functions, there exists an infinite number of solutions\citep{hyvarinen1999nonlinear}. To illustrate, consider that there is an infinite number of possible nonlinear decompositions of a random vector into independent components, and those decompositions are not related to each other in any trivial way. A related problem is that measuring true independence between distributions is often intractable. While ICA can be efficiently solved using approximated independence measures, such as kurtosis, these approaches do not transfer to the nonlinear scenario.

Perhaps the most common approach to solve NICA, which addresses both of the problems, is to pose a constraint on the nonlinear transformation~\citep{lee1997blind, tan2001nonlinear,almeida2003misep,almeida2004linear,dinh2014nice}. One of the first attempts was to generalize ICA by introducing nonlinear mixing models in which case the solution is still possible to identify~\citep{lee1997blind}. In \citep{le2011ica} authors propose Reconstruction ICA (RICA) which requires that mixing matrix $W$ is as close as possible to orthonormal one $WW^T = I$. Thanks to such constraints, one can directly apply independent measures from classical ICA method. 

The aforementioned approaches are arguably limited in their expressive power. In a more recent attempt \citep{dinh2014nice} the authors propose a neural model for modeling densities called Nonlinear Independent Component Estimation (NICE). The authors parameterize the neural network so that it is fully invertible and the output distribution is fully factorized (independent). However, the model incorporates learning the unmixing function using  maximum likelihood, which requires specifying a prior density family. 

Our work is most closely related to the recently introduced Adversarial Non-linear ICA model (ANICA) \citep{brakel2017learning}. In contrast to the previous methods, ANICA does not make any strong explicit assumptions on the transformation function. Instead, a clever adversarial-based measure for estimating and optimizing independence efficiently is proposed. In this work we will take a closer look at this measure, and argue that the basic premise permits construction of an effective non-parametrized independence measure. 

Finally, let us note that a large process has been made in learning factorized representations using deep neural networks~\citep{DBLP:journals/corr/abs-1804-03599,NIPS2016_6399}. What separates ANICA and our method from the previous work is the direct encouragement of independence in the latent space.  A similar path was also taken by~\citep{2018arXiv180205983K} where the VAE loss function is augmented with a cost term directly encouraging disentanglement.

\section{Independence measure by Cramer-Wold distance}\label{sec:theo}

In this chapter we develop an efficient independence measure, which contrary to ANICA model, does not require adversarial training. Our approach can be effectively used to solve nonlinear ICA, in contrast to many other metrics used solely in the context of linear ICA.  

In the following we discuss three independence metrics. Firstly, we  consider  distance correlation, and adversarial--based metric used in ANICA. In the last part we introduce our Cramer-Wold based independence metric.

\paragraph{Distance correlation} 
One of the most well-known measures of independence of random vectors $\X$ and $\Y$ is the {\em distance correlation} (dCor) \citep{szekely2007measuring}, which is applied in \citep{matteson2017independent} to solve the linear ICA problem. Importantly, $dCor(\X,\Y)$ equals zero if the random vectors $\X$ and $\Y$ are independent. Moreover, $dCor$ has a closed-form estimator.

However, to ensure the independence of components of a given random vector $\X$ in $\R^D$, one has to compute $dCor(\X_J,\X_{J'})$ for every subset\footnote{Except for the trivial cases when either $J$ or $J'$ is emptyset.} of indexes $J \subset \{1,\ldots,D\}$, where $J'$ denotes the complement set of $J$ and $\X_J$ is the restriction of $\X$ to the set of coordinates given by $J$. As this procedure has exponential complexity with respect to the number of dimensions, we decided to use a simplified version of $dCor$ which enforces only pairwise independence of the components:
$$
dCor_{pairwise}(\X)=\sum_{i<j} dCor(\X_i,\X_j), 
$$
where $\X=(\X_1,\ldots,\X_D)$.

\paragraph{Adversarial--based independence metric} Now let us describe the adversarial approach used in ANICA. The basic idea  is to leverage that a random permutation of features in a sample produces samples that come from a distribution with independent components. More precisely, let $\X$ be a random vector which comes from pdf $f(x_1,\ldots,x_D)$, and let $X=(x_i)_{i=1..n} \subset \R^D$ be a sample from $\X$, where $x_i=(x_i^1,\ldots,x_i^D)$. We will describe how to draw a sample from the density
$$
F(x_1,\ldots,x_D)=f_1(x_1) \cdot \ldots \cdot f_D(x_D),
$$
where $f_i$ are the marginal densities of $f$. To do this, simply randomly choose maps $\sigma_i$ from $\{1,\ldots,n\}$ into itself, and consider 

$$
X_{shift}:=(y_i)_{i=1..n}, \mbox{ where  }
y_i=(x_{\sigma_1(i)}^1,\ldots,x_{\sigma_D(i)}^D).
$$
Then $X_{shift}$ comes from the pdf $F$, which has independent components. Consequently, if $X$ and
$X_{shift}$ are close, then the same holds for $f$ and $F$, and consequently $f$ has independent components. In ANICA adversarial training is used to reduce distance between
 $X$ and $X_\sigma$.

\paragraph{Cramer-Wold independence metric} 
The application of adversarial training in ANICA can lead to instability, as discussed by the authors, and slower training. In this paper we propose an alternative independence measure.
Our main idea is to compute the distance between $X$ and $X_\sigma$ without resorting to adversarial training.

In order to achieve this, one can choose commonly used metrics, such as the Kullback-Leibler divergence \citep{kingma2014auto} or Wasserstein distance \citep{tolstikhin2017wasserstein}. Instead, due to its simplicity, we have decided to use the recently introduced Cramer-Wold distance $d_{\CW}$ \citep{tabor2018cramer}, which also possesses the advantage of having the closed-form for the distance of two samples\footnote{In the computation we apply the equality $\phi_D(0)=0$.} $X=(x_i)_{i=1..n},Y=(y_i)_{i=1..n} \subset \R^D$:

$$
\begin{array}{l}
d^2_{\CW}(X,Y)\\
=\frac{1}{2n^2\sqrt{\pi \gamma}}
\big(\sum \limits_{ii'}\phi_D(\frac{\|x_i-x_{i'}\|^2}{4\gamma}) \\
\phantom{\frac{1}{2n^2}}
+\sum \limits_{jj'}\phi_D(\frac{\|y_j-y_{j'}\|^2}{4\gamma})-2\sum \limits_{ij} \phi_D(\frac{\|x_i-y_j\|^2}{4\gamma}) \big).
\end{array}
$$

where the  bandwidth $ \gamma$ is a hyperparameter, which may be set accordingly to the one-dimensional Silverman's rule of thumb to $ \gamma =(\frac{4}{3n})^{1/5}$. The function  $\phi_D$ is computed with the asymptotic formula:~$\phi_D(s) \approx (1+\frac{2s}{D})^{-1/2}$.

As a final step, we normalize each component of  $X_{shift}$ to ensure that the Silverman's rule of thumb is optimal, and define our independence metric as: 
\begin{equation} \label{eq:ii}
ii_D(X):=d^2_{\CW}(X,cn (X_{shift})),
\end{equation} where $cn(Y)$ is the componentwise normalization of $Y$.

\begin{figure*}[!h]
    \centering
 \includegraphics[width=0.31\textwidth]{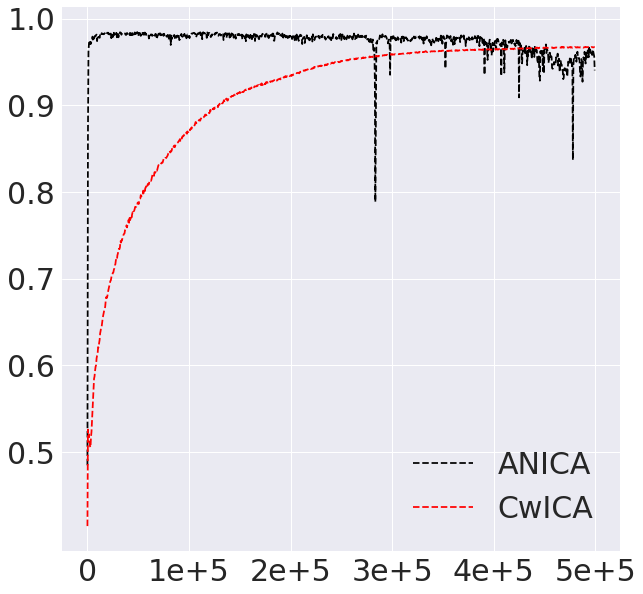} 
 \includegraphics[width=0.31\textwidth]{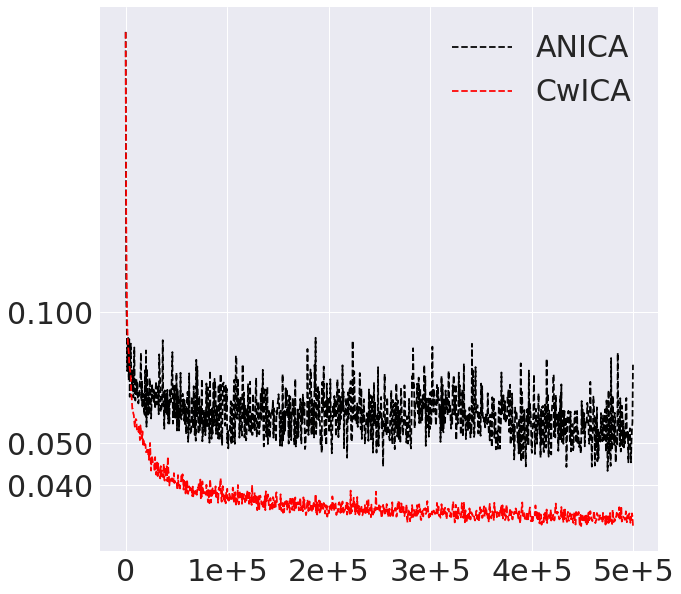} 
  \includegraphics[width=0.31\textwidth]{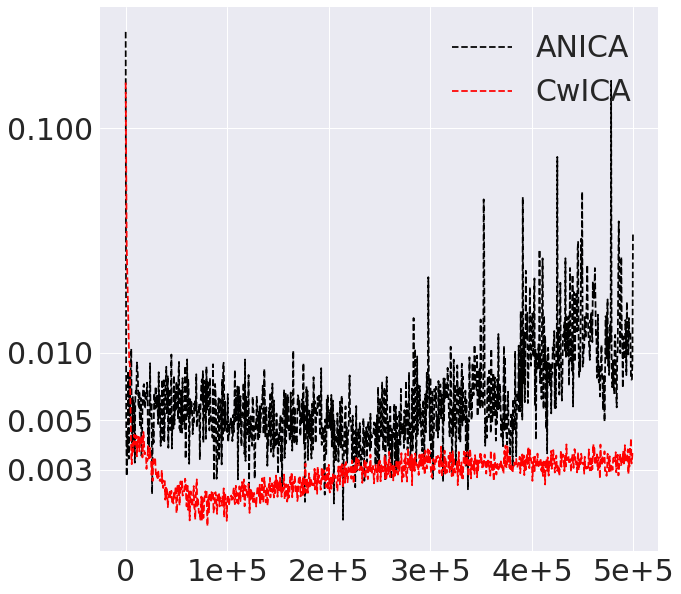} 
    \caption{The number of iterations versus $max\ corr$ (left), $MSE$ (middle) and $dCor$(right) for ANICA (black) and CwICA (red) \ola{} Please note that the $MSE$ and $dCor$ results are plotted in logarithmic scale on the y-axis. This experiment is separate to the one presented in table \ref{table:syn_nonlin}, therefore the results may slightly differ.  }
    \label{fig:compare_CW_ANICA}
\end{figure*}

\section{Algorithm}
\label{sec:alg}

We are now ready to define \CWICA, a nonlinear ICA model based on the Cramer-World independence metric. Following ANICA, we use an Auto-Encoder (AE) architecture.

Let $X \subset \R^N$ denote the input data. An Auto-Encoder is a model consisting of an encoder function $\E:\R^N \to \Z$ and a complementary decoder function $\D:\Z \to \R^N$, aiming to enforce coding of the input variables that minimizes  the reconstruction error:
\begin{equation}
\recerror(X;\E,\D)=\sum_{i=1}^n \|x_i-\D(\E x_i)\|^2.
\end{equation}

The goal of our method is to train an encoder network $\E X$ which maps data to informative, statistically independent features $Z$. In order to achieve this we introduce an independence measure on the latent space, by taking advantage of the independence index $\ii_D(\E X)$ defined in (\ref{eq:ii}). We denote this model as the \CWICA (Cramer-Wold Independent Component Analysis).  

To obtain a procedure independent of a possible rescaling of the data, we have decided to use a multiplicative model instead of an additive:
\begin{equation} \label{eq:cost}
\mathrm{cost}(X;\E,D)=\ii_D(\E X) \cdot \recerror(X;\E,\D).
\end{equation}

\begin{figure*}
   \centering
\includegraphics[width=0.46\textwidth]{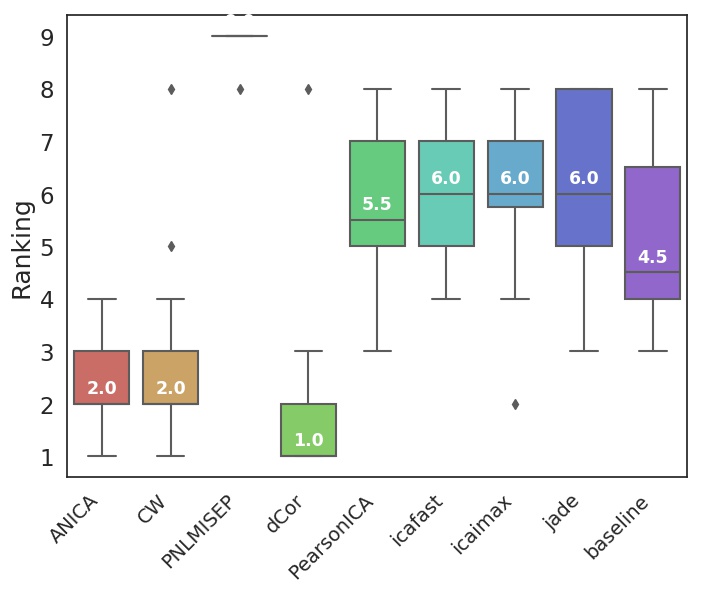} \includegraphics[width=0.46\textwidth]{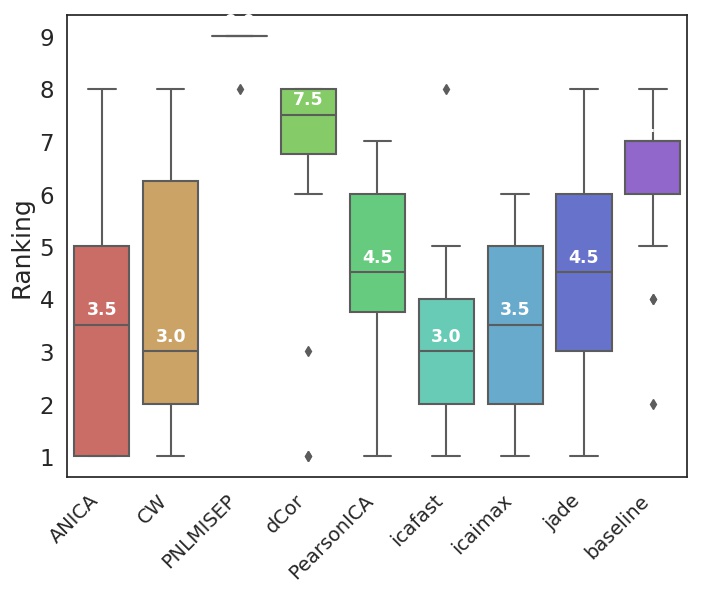} 
   \caption{The ranking (lower is better) of algorithms based on mean maximum correlation between the latent variables and sources (left-hand side) and dCor (right-hand side) in dimension $d=10$. }
   \label{fig:ranking_10}
\end{figure*}

In contrary to ANICA we do not use an adversarial
objective, proposing instead a close-form solution based on the independence index.  However, enforcing independence by itself does not guarantee that the mapping from the observed signals $X$ to the predicted sources $Z$ is informative about the input. Therefore, the decoder constrains the encoder, as proposed in ANICA. 

As explained earlier, in the case of Cramer Wold index it is important to normalize the resampled  (permuted) latents, which additionally prevents the encoder's output from vanishing or exploding in magnitude. 

\begin{algorithm}
\algsetup{linenosize=\tiny}
\small
\caption{(\bf CwICA train loop:)  }
\label{allg1}
\begin{algorithmic}
			\STATE {\bf input}
			\STATE\hspace\algorithmicindent data $\bf X \in \R^d$, with each sample in a separate row
			\STATE\hspace\algorithmicindent encoder $\E$, decoder $\D$
				\REPEAT 
                \STATE \textit{sample} a batch $X$ of size $n$ from $\bf X$
                \STATE \textit{apply encoder} $Z = \E X $
                \STATE \textit{resample to obtain $Z_{shift}$:}
                \FOR{$i \in \{ 1, \ldots, n \}$ }		
							\FOR{$ j \in \{ 1, \ldots, d \}$ }				
								\STATE $k \sim Uniform(\{1,\ldots, d\})$ // sample col. index
								\STATE ${Z_{shift}}_{i,j} = Z_{i,k} $
						\ENDFOR
                \ENDFOR	
                \STATE \textit{normalize} ${Z_{shift}}$ by element-wise rescaling 
                $$
                 {  { \hat Z_{shift_{\cdot, j}}} } = \frac{{Z_{shift}}_{\cdot, j} - \mean({Z_{shift}}_{\cdot, j})}{\std({Z_{shift}}_{\cdot, j}) } \mbox{\quad for  } j=1,\ldots d 
                $$                 
                \STATE 
                    $
                    J = d^2_{\CW}(  { \hat Z_{shift}} , Z  )  \cdot \recerror(X;\E,\D)
                    $
			    \STATE Update $E$ and $D$ to minimize $J$
				\UNTIL{ converged}

\end{algorithmic}		
\end{algorithm}

\begin{table*}[!t]
\centering
\scalebox{1.0}{ 
\begin{tabular}{ c | c | c | c| c | c | c | c |c}
&	ANICA &	CwICA & PNLMISEP	 & dCorICA & PearsonICA & icafast exp & icaimax ext & jade \\
\hline
$dCor$ 	& 0.0027 &	0.0017 & --- & \bf0.0000 & 0.0017	& 0.0017&0.0017	&0.0017\\
$max\ corr$	& \bf0.9835 &	0.9697 & --- & 0.3033 &0.8969	&0.8926	&0.8940	&0.9414 \\
$MSE$	&0.0516	& \bf0.0332	& --- &0.1475 & --- &---&---&---\\
\end{tabular}
}
\caption{Results on nonlinear synthetic data}
\label{table:syn_nonlin}
\end{table*}
\begin{table*}[!t]

\centering
\scalebox{1.0}{ 
\begin{tabular}{ c | c | c | c | c | c|c|c|c}
 &	ANICA &	CwICA & PNLMISEP &	dCorICA  & PearsonICA & icafast exp & icaimax ext & jade \\
\hline
$dCor$	& 0.0027 & 0.0175 &0.0080 & \bf 0.0000 & 0.0038&	0.0038&	0.0038&	0.0038 \\
$max\ corr$	& 0.8913 & 0.7805 &  0.9012 & 0.2514 & 0.9997 &	0.9997 & \bf 0.9998 &	0.9984 \\
$MSE$	& 0.0333 & \bf 0.0094 & --- & 0.1746 & --- & --- & --- & ---\\
\end{tabular}
}
\caption{Results on linear synthetic data}
\label{table:syn_lin}
\end{table*}

\begin{figure*}[!h]
    \centering
 \includegraphics[width=0.90\textwidth]{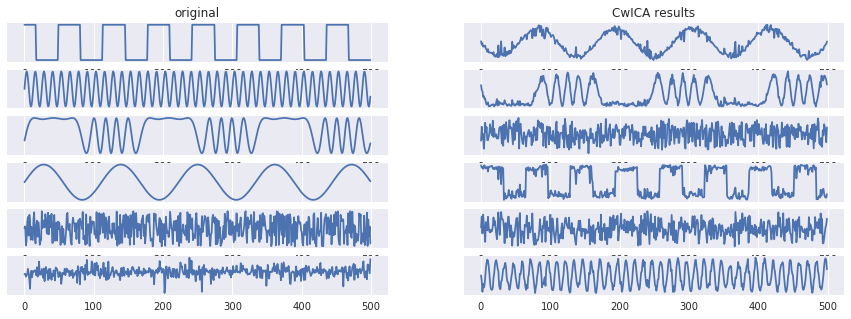} 
    \caption{The original sources (left) and the independent components predicted by CwICA (right) obtained from nonlinear mixtures.}
    \label{fig:syn_res_nonlin}
\end{figure*}

\begin{table*}[!t]
\centering
\scalebox{1.0}{ 
\begin{tabular}{ c |  c c c c | c c c c }
& \multicolumn{4}{c}{$max\ corr$} & \multicolumn{4}{c}{ $dCor$ } \\ 

&  2 & 5 & 10 & 20  &  2 & 5 & 10 & 20 \\

\hline

ANICA&0.78&0.67&0.69&\bf 0.7    &\bf{0.17}&\bf{0.13}&\bf{0.10}&0.14 \\
CwICA&\bf 0.79& 0.69&0.66&0.68	&0.22&0.19&0.15&0.12 \\
PNLMISEP&0.77&\bf0.71& -- & --	&0.18&0.15& -- & -- \\
dCorICA&\bf 0.79&0.68&\bf 0.73&0.67 &0.24&0.20&0.20&0.23 \\
\hline
PearsonICA&0.73&0.61&0.59&0.57  & 0.29& 0.21&0.12&0.10 \\
icafast &0.75&0.59&0.59&0.57 &0.21&0.19&\bf{0.10}&\bf{0.09} \\
icaimax &0.75&0.60&0.59&0.57 &0.21&0.19&\bf{0.10}&\bf{0.09} \\
jade&0.74&0.59&0.59&0.57 &0.25&0.20&0.11&0.10 \\
\hline
baseline&0.70&0.60&0.61&0.59 &0.36&0.24&0.15&0.11 \\

\end{tabular}
}
\caption{Results on nonlinear image dataset. For dimension $10$ and $20$ the PNLMISEP did not converge. }
\label{table:img_nonlin}
\end{table*}

In addition we implement another AE-based, nonlinear model, which follows the same architecture as CwICA, but substitutes $ii_D(\E X)$ by $dCor(\hat Z )$. From this point onwards and in all figures and tables, for simplicity we shall also use the $dCor$ notation instead of the $dCor_{pairwise}$. The $\hat Z$ stands for the component-wise normalized features of the encodings of $X$. We refer to this method as $dCorICA$. 

\section{Experiments} \label{se:ex}

We evaluate our method on mixed images and synthetic dataset. For comparison we use the nonlinear method ANICA \citep{brakel2017learning} and the PNLMISEP \citep{Zheng2007PNLMISEP}, an extension to the MISEP method \citep{almeida2003misep,almeida2004linear}. It should be noted that the PNLMISEP is designed especially for post-nonlinearity, not for the more general nonlinear mixing functions used in presented experiments. We also report the results obtained on the same datasets by four selected linear models. We choose the popular FastICA algorithm \citep{hyvarinen1999fast}, the Information-Maximization (Infomax) approach \citep{bell1995information},  the Joint Approximate Diagonalization of Eigenmatrices (JADE) \citep{cardoso1993blind} and the Pearson \citep{stuart1968advanced} system PearsonICA \citep{karvanen2000pearson}. We use the implementations of the linear models in R packages {\tt ica} \citep{ica} and {\tt PearsonICA} \citep{pearsonica}.

\subsection{Comparison with ANICA}

The CwICA and dCorICA models follow a similar architecture as ANICA, but use a closed form independence measure on the latent variables, as opposed to the adversarial approach. We compare our algorithms with the ANICA model using the synthetic signals dataset defined in \citep{brakel2017learning}.

The dataset in the nonlinear setting consists of $n=4000$ observations $X \in \R^{n \times 24}$ which are obtained by applying mixing function $X=tanh(tanh(YA)B)$ to the independent sources $Y \in R^{n \times 6}$, where $A$ and $B$ are sampled uniformly from $[-2,2]$ and $tanh$ is the hyperbolic tangent function. We select the first $500$ samples as the test dataset, and train on the remaining $3500$ samples. We fit ANICA using the best hyper-parameters setting for this dataset reported by \citep{brakel2017learning}. For  CwICA we perform a grid search on the learning rate and bandwidth, using batches of size $256$ and choose the model with the smallest total loss on the validation dataset. The validation dataset has size $500$ and is drawn from the same distribution as the train and test sets. All other model hyper-parameters are set as in ANICA. We also ran a similar grid search on learning rate and batch size for dCorICA.
We do not execute the PNLMISEP, as the implementation of this method is not suitable for input data of this dimensionality. 

We also report the performance of the nonlinear methods on linear data. The linear dataset is obtained from the same independent sources $Y$ by a transformation defined by the matrix $A$. We train the models using the same configuration as in the nonlinear experiment.

\begin{table*}[!t]
\centering
\scalebox{1.0}{ 
\begin{tabular}{ c |  c c c c | c c c c }
& \multicolumn{4}{c}{$max\ corr$} & \multicolumn{4}{c}{ $dCor$ } \\ 

&  2 & 5 & 10 & 20  &  2 & 5 & 10 & 20 \\

\hline
ANICA&0.90&0.74&0.73&0.7 &0.16&0.11&\bf0.09&0.14 \\
CwICA&0.85&0.73&0.74&0.68 &0.23& 0.15&\ 0.11&0.10 \\
PNLMISEP&0.87&0.74& -- & -- &\bf0.14&\bf0.09& -- & -- \\
dCorICA&0.89&0.74&0.76&0.57  &0.30& 0.15& 0.11&  0.28 \\
\hline
PearsonICA&0.91&0.82&0.8&0.67   & 0.25&0.14& 0.11&0.18 \\
icafast &0.92&0.83& \bf  0.82&0.75   &0.22& 0.15&0.10&\bf0.08 \\
icaimax &0.91& \bf 0.84&\bf 0.82&\bf 0.77   &0.24&0.14&0.10&0.10 \\
jade&\bf 0.93& \bf  0.84&0.79&0.68  &0.23&0.14&0.10&0.09 \\
\hline
baseline&0.85&0.72&0.71&0.65  &0.33&0.16&0.11&0.09 \\
\end{tabular}
}
\caption{Results on linear image dataset.  For dimension $10$ and $20$ the PNLMISEP did not converge.}
\label{table:img_lin}
\end{table*}

We evaluate the methods on test data using the mean $dCor$ distance between all possible pairs of the unraveled latent independent factors $Z$. In addition, we compute the mean maximum correlation (denoted as $max\ corr$) between the sources $Y$ and the results $Z$. As ICA extracts the source signals only up to a permutation, we consider all possible pairings of the predicted signals with the source signals and report only the highest $max\ corr$ value. Before computing the $dCor$, the latents $Z$ are normalized. The results are presented in tables \ref{table:syn_nonlin} and \ref{table:syn_lin}. The original sources and the recovered by CwICA signals are presented in figure \ref{fig:syn_res_nonlin}.

\begin{table}[!t]
\centering
\scalebox{1.0}{ 
\begin{tabular}{ c | c | c | c }
dim & ANICA & CwICA & dCorICA \\
\hline
2 & 0.5839 & \bf0.0097 & 0.6041 \\
5 & 0.5811 & \bf0.0181 & 0.5491 \\
10 & 0.5146 & \bf0.0389 & 0.4616 \\
20 & 0.5299 & \bf0.2748 & 0.5079 \\
\end{tabular}
}
\caption{Reconstruction loss (MSE) for auto-encoders on the nonlinear image dataset.}
\label{table:img_rec}
\end{table}

CwICA behaves very well on the nonlinear dataset, achieving a similar $max\ corr$  value to ANICA, at the same time outperforming it in $MSE$ and $dCor$ criterions. This makes the method the best choice if a balanced solution is desired.

In addition we run the ANICA and CwICA models $5$ times with different seeds. We pick the best model in terms of $dCor$ and summarize the reported metrics on the validation dataset during training in figure \ref{fig:compare_CW_ANICA}. In this experiment both models where trained using batch size $256$.

In the linear synthetic data experiments all non-linear models perform worse than the classical ICA algorithms. This sustains the claim that if the linear characteristic of the mixing function is assumed beforehand, the most efficient is the use of dedicated methods.

The dCorICA algorithm, as expected, achieves the lowest $dCor$ cost in both linear and nonlinear setting; however, fails to recover the original sources. This may suggest that the model focuses on the minimization of the independence loss, disregarding the information in the input.

\subsection{Comparison on image dataset}

One of the most popular applications of ICA is separation of images. We conduct experiments on a dataset composed of images from the USC-SIPI Image Database, scaled to $100 \times 67$ pixels and mixed using $X=f(tanh(YA)B)$, where $Y$ are the original sources, $X$ are the observations, $dim(Y)=dim(X)$, $dim(X) \in \{2,5,10,20\}$, $f(x)=x^2+x^3$ is applied element-wise, and $A$ and $B$ are sampled uniformly from interval $[-2,2]$. In addition, we prepare a linear dataset, where the mixing function is defined by the transformation imposed by a random matrix $C$ sampled uniformly from $[-2,2]$.  The components of $Y$ are separate, flattened, gray-scale images, chosen at random from a dataset of size $100$. The observations $X$ are normalized before passing to the algorithms.  The numbers of distinct observation examples for each dimension are $50, 50, 20,10$, respectively.

For each $dim(X)$ we test the ANICA, CW, dCor, PNLMISEP, FastICA, Infomax, Jade and  PearsonICA algorithms.  All the nonlinear models are trained using the same configurations as in the previous subsection. We report the mean $max\ corr$ and  $dCor$ distance for each method  in Table \ref{table:img_nonlin}  (nonlinearly mixed data) and in Table \ref{table:img_lin}  (linearly mixed data). We also report the $MSE$ loss for auto-encoders (ANICA, CW, dCor) in Table \ref{table:img_rec}.

\CWICA{} achieves high $max\ corr$ on the nonlinearly mixed data, comparable to the other non-linear ICA algorithms (in fact CwICA gets the best results among all ICA algorithms for $dim(X) = 2$) and strongly outperforms ANICA and dCorICA separations on reconstruction loss.

Additionally, dCorICA gives satisfactory results on the nonlinear setting only for low dimensional data ($dim(X) \in \{2,5\}$). For $dim(X) \ge 10$  dCorICA still manages to compete with other models in $max corr$, but evidently obtains the worst results in $dCor$, despite the fact that it minimizes this measure directly. This disproportion can be especially observed in Figure \ref{fig:ranking_10}, which presents the mean rank of the methods based on the two metrics.

For higher dimensions, the nonlinear methods perform better in $max corr$; however, fail to surpass the classical algorithms in terms of $dCor$.  An opposite  trend in the linear data experiments may be observed for the lower dimensions (up to $10$). In general, the linear methods achieve much better  $max_{corr}$, and worse (higher)  $dCor$.  For $dim(X)=20$ in both nonlinear and linear setting, the results obtained by auto-encoders  are even worse than the baseline scores.

\section{Conclusions}

In this paper we have proposed a closed-form independence measure and applied it to the problem of nonlinear ICA. The resulting model, CwICA, achieves comparable results to ANICA, while by using a closed-form formula avoids the pitfalls of adversarial training. Future work could focus on scaling up these approaches to higher dimensional datasets, and applying the developed independence metric in other contexts. Finally, we found that nonlinear methods generally under--perform on linearly mixed signals, which could be addressed in future work.

\newpage

\bibliographystyle{named}
\bibliography{ijcai2019}

\end{document}